%% file: Main.tex
\documentclass[a4paper]{report}
\usepackage[utf8]{inputenc}
\usepackage[T1]{fontenc}
\usepackage{RJournal}
\usepackage{amsmath,amssymb,array}
\usepackage{booktabs}

\usepackage{hyperref}
\usepackage{verbatim}
\usepackage[font={bf}]{caption}
\usepackage{listings}
\usepackage{color}
\usepackage{algpseudocode}
\usepackage{subcaption}
\usepackage[Algorithmus]{algorithm}
\usepackage{multirow}

\definecolor{codegreen}{rgb}{0,0.6,0}
\definecolor{codegray}{rgb}{0.5,0.5,0.5}
\definecolor{codepurple}{rgb}{0.58,0,0.82}
\definecolor{backcolour}{rgb}{0.95,0.95,0.92}
\definecolor{ashgrey}{rgb}{0.7, 0.75, 0.71}
\definecolor{airforceblue}{rgb}{0.36, 0.54, 0.66}
 
\lstdefinestyle{mystyle}{
    backgroundcolor=\color{backcolour},   
    commentstyle=\color{codegreen},
    keywordstyle=\color{airforceblue},
    numberstyle=\tiny\color{codegray},
    stringstyle=\color{codepurple},
    basicstyle=\footnotesize,
    breakatwhitespace=false,         
    breaklines=true,                 
    captionpos=t,                    
    keepspaces=true,                 
    numbers=left,                    
    numbersep=5pt,                  
    showspaces=false,                
    showstringspaces=false,
    showtabs=false,                  
    tabsize=2
}
 
\begin{document}

\sectionhead{Preprint research article}
\volume{XX}
\volnumber{YY}
\year{20ZZ}
\month{AAAA}

\begin{article}
  \input{prestesgarcia}

\end{article}

\end{document}

%% file: prestesgarcia.tex
\title{Applying Evolutionary Metaheuristics for Parameter Estimation of Individual-Based Models}

\author{by Antonio Prestes Garc\'ia, Alfonso Rodr\'iguez-Pat\'on}
\maketitle

\abstract{
Individual-based models are complex and they have usually an elevated number of input parameters which must be tuned for reproducing the observed population data or the experimental results as accurately as possible. Thus, one of the weakest points of this modelling approach lies on the fact that rarely the modeler has the enough information about the correct values or even the acceptable range for the input parameters. Consequently, several parameter combinations must be tried to find an acceptable set of input factors minimizing the deviations of simulated and the reference dataset. In practice, most of times, it is computationally unfeasible to traverse the complete search space trying all every possible combination to find the best of set of parameters. That is precisely an instance of a combinatorial problem which is suitable for being solved by metaheuristics and evolutionary computation techniques. In this work, we introduce EvoPER, an R package for simplifying the parameter estimation using evolutionary computation methods.
}

\section{Introduction}

Modeling and simulation is certainly a vast discipline with a broad and complex body of knowledge having, beyond the surface, a large technical and theoretical background \citep{Minsky1965} \citep{Banks2009} \citep{citeulike:191333} \citep{citeulike:606462} which consequently, is hard of being completely mastered from modelers coming from disperse domains like biology, ecology or even computer science. Among the existing formalisms, the agent-based or individual-based is increasing gradually the number of adepts in the recent years. The Individual-based modeling is a powerful methodology which is having more and more acceptance between researchers and practitioners of distinct branches from social to biological sciences, including specifically the modeling of ecological processes and microbial consortia studies. Certainly, one of the main reasons for the success of this approach is the relative simplicity for capturing micro-level properties, stochasticity and spatially complex phenomena without the requirement of a high level of mathematical background \citep{GrimmRailsback2005}. But the counterpart of the ease for building complex and feature rich models, is the lack of a closed formal mathematical form of the model which implies that the study of these models cannot be attacked analytically. Thereby, the only way to explore and adjust the parameters of these type of models is the brute-force approach, executing the model many and many times and evaluating the results of each execution. 

The systematic search of for the best set of model parameters is a costly task for which there are basically two different types of approaches for exploring the solution space of simulation outputs. The first one is using some static sampling scheme based on the design of experiments (DoE), such as randomized, factorial or Latin hypercube \citep{Little1978} \cite{Loh1996} \citep{Thiele2014} designs which works by generating a collection of sampling points of parameter space which are further evaluated. Alternatively, the parameter estimation can be also stated as an instance of an optimization problem and therefore, addressed more conveniently using the whole arsenal of metaheuristics and evolutionary strategies. The main difference between these two ways of tackling with the problem lies on the fact that the first one is fundamentally a static sampling technique whereas the second is an intrinsically dynamic form of a guided partial search over the solution space where the set of initial solutions are continuously improved and hopefully converging towards a global optimum \citep{Weise2009} \citep{Boussaid2013}.  It turns out that comparatively, the optimization approach may require less model evaluations to find the best, or at least an acceptable solution for the parameter estimation problem which, in the case of models with an average complexity level, means a difference between an upper bound computational cost from hours to days. 

The parameter estimation of individual-based models is a particularly hard instance of an optimization problem as they are highly stochastic and they parameter space most often tend to show nonlinear interactions which difficult the localization of good parameter combinations producing the minimal deviation from reference data. Differently from deterministic and closed form models, where the most significant computational cost is due to the optimization algorithm itself, in the case of individual-based models the time required every single model execution alone is responsible for most of the computational time taken in the optimization process.  It is important to consider that some metaheuristics may produce better results than others depending on the problem type and more specially on the structure and characteristics of model being analyzed. Therefore, it is interesting before undertake a full length run to explore different algorithms in order to find the best suited for the problem instance.

In the next sections, we will briefly describe the scope of parameter estimation problem and the usage examples of the EvoPER R package which has been developed for facilitating the tasks of estimating the parameters of Individuals-based models. The current version of EvoPER includes implementations of Particle Swarm Optimization (PSO), Simulated annealing (SA) and Ant Colony Optimization (ACO) algorithms developed exclusively for this work and adapted for their use in the parameter estimation of agent-based models. We are also introducing two simple evolutionary strategies implemented for exploratory analyses for the parameter space of individual-based models which can be useful for mapping the promising zones of solution space.


\section{Parameter Estimation and Optimization}

The terms model calibration and parameter estimation, although informally are used interchangeably and being functionally similar are semantically distinct entities having a different scope and objectives. Therefore, to provide a more formal definition of these terms let us briefly define the basic structure of a mathematical model. A model is normally expressed as some form of the algebraic composition expressing the relationship between of three element types, namely the independent variables, the dependent or the state variables and finally the constants. For the sake of simplicity, a model expressing some linear relationship between variables is shown bellow

$$y = \alpha + \beta x$$

where $x$ and $y$ are independent and the state variable respectively and $\alpha$ and $\beta$ are the model constants. More generally, the structure of and stochastic model can be represented by the functional relation \cite{Haefner} given by the expression shown in the Equation ~\eqref{eq:model1},

\
\begin{equation}
\label{eq:model1}
y = f(\vec{x}, \vec{p}, \vec{\epsilon}),
\end{equation}
\

The terms $\vec{x}$, $\vec{p}$ and  $\vec{\epsilon}$ denotes respectively the vectors of independent variables, the vector of model parameters and the stochastic deviations.

The model constants are referred as the model parameters which necessarily do not have to have any correspondence to some element in the system being modeled \citep{PE.Beck1977}. The direct problem is, being known the model structure and, also knowing the independent variables and the parameters, to estimate the value of state variable. Of course, this oversimplified case is rarely seen when modeling real systems, especially when dealing with biological systems. In addition, in the most cases the constants and the independent variables are impossible to observe directly being also unknown the right model structure for representing the system under study.

Usually the only value elucidated experimentally or backed by observations of some population data is the state variable; therefore, the parameters which are the structural part of model must be estimated having as the only reference, the measurements of dependent variable. Hence the term {\bf calibration} can be defined as the procedure to where the values of state variable $"y"$ are compared to the known standard values, let's say $"Y"$, which in the context of biological research are those sampled from population true values \citep{citeulike:191333}. 

On the other hand, the {\bf parameter estimation} is the task of estimating the values of the constants of a model and it can be seen somehow as an inverse problem, since we are using the reference values $Y$ in order to determine the suitable values for the model constants \citep{Ashyraliyev2009} \citep{PE.Beck1977}. The parameter estimation procedure implicitly encompasses the calibration process as, in order to discover the values for the constants the model outputs must be checked to the reference values. Thus the problem can be also stated as an optimization problem, just because the process requires the search for the minimum values of some function $f(y_i, Y_i)$ measuring the distance between $y_i$ and  $Y_i$ which are the simulated and the reference values respectively.

The family of functions measuring how close are the simulated and the reference values is the goodness of fit metric of a model and is known as the objective function. The objective function facilitates the determination of how well the model is able to reproduce the reference data. In other words, the objective function provides a numerical hint about how close are the output of model to the reference data.  For any given model, a family of different objective functions can be defined over the output data, depending on the chosen distance metric and on what is the target of parameter estimation process. More formally, the objective function can be defined as $ f \colon \mathbb{R}^n \to \mathbb{R} $ and can be further generalized for an individual-based model where, differently from a pure mathematical form, the objective function domain may assume any valid computational type. Thus, being $\mathbb{S}$ the set of valid computational model parameters, the objective function can be rewritten as  $ f \colon \mathbb{S} \to \mathbb{R} $.

Therefore, every candidate solution $x$ is instantiated from the solution set $\mathbb{S}$ being the best of them known as the solution or the optimal and represented as $x^*$. Hence, the target of optimization process is to find the solution $x^*$ which minimize the objective function such that $f(x^*) < f(\mathbb{X}) $, being $\mathbb{X}$ the set of all candidate solutions. It is worth to mention that although uncommon, the optimization can be also defined as a maximization process. Another important aspect to note is that the objective function can be much more than a simple distance measurement, thus more complex tasks can be carried out using an algorithmic approach \cite{Weise2009}. We will illustrate that kind of approach, tuning a problem solution for making model output to oscillate in a fixed period in the example section of this work.

There are fundamentally three approaches to define the distance metrics for a model \citep{Thiele2014}. The first approach is based on using acceptable ranges for the model outputs being the most straightforward one. That approach is also known as categorical calibration and works defining intervals for the model output values and when the output falls inside the interval it is considered as having a good fit. One of the main drawback of this approach is the fact that it is not possible to determine how close are the model and the reference data. The second metric relies on measuring the differences between simulated and observed values, being the least squares the most commonly used method for computing the quality of fit \citep{PE.Beck1977}. Finally, that last approach requires the use of likelihood functions. It is hard to implement and requires that the underlying distribution must be known, which usually precludes its application on complex non-linear computer models. 

The systematic exploration of solution space which is compulsory for the calibration process requires many model executions as well as many evaluations of goodness of fit function over the output data to find the best estimation for the model parameters provided that they minimize the discrepancies between simulated and observed values. This is a computationally expensive task, especially in the case of Individual-based models, as the problem bounds increases with model complexity and the number of input parameters which must be tested. Roughly speaking there are basically two different approaches for generating the sampling points required for estimating parameters. The first of them is based on the definition of sampling schemes such as Monte Carlo sampling, Factorial designs or the Latin Hypercube sampling. These techniques work by generating an a priori set of samples in the search space, that is to say, a collection of parameter combinations which are further used for running model and evaluating the cost function \citep{Thiele2014} \citep{Viana2013}.  The Latin hypercube sampling is a generalization of the Latin squares classical experimentation design randomization typically found on agricultural experimentation \cite{Little1978}.

On the other hand, in the case of using optimization methods, only an initial set of points, sampled from the input space are instantiated and these solutions are updated dynamically searching for neighboring solutions which could approximate better towards to the minima. The exact implementation details, depends on the metaheuristic chosen for the parameter estimation process, but despite of diversity of existing metaheuristics practically all of them can be functionally described and completely characterized by combining the building blocks contained in the pseudocode shown in Figure \ref{fig:EA-1}.

\begin{figure}[h]
\begin{lstlisting}[mathescape, language=Pascal]
$P_0 \leftarrow $ Initialize()
$ f \leftarrow $ Evaluate($P_0$)
while !terminate
	$P^{\prime} \leftarrow $ Selection($P_0, f$)
	$P^{\prime} \leftarrow $ Recombination($P^{\prime}$)
	$P^{\prime} \leftarrow $ Mutation($P^{\prime}$)
	$ f^{\prime} \leftarrow $ Evaluate($P^{\prime}$)
	$(P_0, f) \leftarrow $ Replace($P_0, f, P^{\prime}, f^{\prime}$)
end	
\end{lstlisting}
\caption[Evolutionary Strategy]{\label{fig:EA-1} The general outline of a metaheuristic optimization method. The pseudocode shows the initialization step followed by the main loop where the initial solution is improved and guided through the search space using the value of the fitness or the solution cost.  }
\end{figure}

Despite of the multiplicity of existing metaphors inspired on a many different sources, ranging from physics or the collective behavior of some eusocial insects to the musical theory \citep{ Sorensen2015}, most of them are just slight variations over the basic evolutionary strategy skeleton. As can be seen in Figure \ref{fig:EA-1} the metaheuristic structure contains few operators depicted in the algorithm by the functions $Selection()$, $Recombination()$, $Mutation()$ and $Replace()$ the other components are the $Initialization()$ and the $Evaluate()$ function. The {\it selection} function is responsible for picking parents from the current population which will be used later for producing the offspring in the next generation. The selection process can be stochastic or using the fitness metric for selecting breeding individuals. The objective of {\it recombination} process is to mimic, in some extent, the genetic chromosomic recombination and mix together $n$, being $n > 1$ parent solutions for producing an offspring which will the combination solution structure of $n$ parents. The {\it mutation} operator, roughly speaking, is to generate stochastically random changes in the solution structure providing the necessary variability for exploring the problem space. Finally, the {\it replacement} process will select the individuals based on their fitness values from the current solution which will be conserved in the next algorithm iteration returning a tuple  $(P_0, f)$ with the individual solutions and its associated fitness. The accessory functions {\it initialize} and {\it evaluate} are required respectively for instantiating the initial solution and for evaluating the fitness metric for the provided solution set $\mathbb{S}$. With respect to the termination condition for the algorithm, the most commonly used approach is a combination of the convergence criteria and the maximum number of iterations. It is worth to say that not necessarily all these components are required to be present on specific metaheuristics. For instance, the simulated annealing \citep{Kirkpatrick1983} uses a population of size equal to one, therefore the selection and recombination process are superfluous in this metaheuristic.

\section{Metaheuristics for Parameter Estimation}

In order to facilitate the parameter estimation task of Individual-based models we introduce the GNU R \citep{GNUR:Manual} package {\bf EvoPER} - Evolutionary Parameter Estimation for Repast, an open source project intended to facilitate de adoption and application of evolutionary optimization methods and algorithms to the parameter estimation of IBMs developed using the Repast Symphony framework \citep{North2013}. The EvoPER package is released under the MIT license being the binaries available for download from CRAN (\url{ https://cran.r-project.org/web/packages/evoper/}) and the complete source code for the project can be found on GitHub (\url{https://github.com/antonio-pgarcia/evoper}).

The package EvoPER provides implementations of most common and successful metaheuristics algorithms for optimization specially crafted for searching the best combination of input parameters for Individual-based models developed in Repast Simphony. Current version of EvoPER package supports the Particle Swarm Optimization (PSO) \citep{Kennedy1995}, the Simulated Annealing (SA) \citep{Kirkpatrick1983} and the Ant Colony Optimization (ACO) \citep{Dorigo2006} algorithms for parameter estimation. We also plan to support more algorithms in successive versions. The metaheuristic algorithms use bio-inspired, natural or physical system analogy having each of them subtleties making them suitable for different types of problems. Nonetheless, despite of the differences in the chosen natural metaphor all algorithms share an important aspect which is that the search space is traversed downhill but allowing uphill moves to avoid to get trapped in a local optimum far from the global one. 

The basic PSO algorithm uses the idea of particles moving in a multidimensional search space being the direction controlled by the {\it velocity}. The velocity has two components, one towards to the direction of best value of particle $p_i$ and other towards to the best value found in the neighborhood of particle $p_i$ \citep{Kennedy1995} \cite{Poli2007}. The behavior and convergence of the algorithm is controlled by the particle population size and by the $\phi_1$, $\phi_2$ parameters which respectively controls the particle acceleration towards the local and the neighbor best.  The algorithm implementation and the default values for the algorithm parameters follows the guidelines and standard values for the algorithm parameters facilitated by \citep{Clerc2012} which are proved to provide the best results. 

The metaheuristic known as Simulated Annealing uses the idea of a cooling scheme to control how the problem solutions are searched. The algorithm generates an initial solution and then iterates, looking for neighbor solutions, accepting new solutions when they are better than the current solution or with some probability $P$ which is function of current temperature and the cost of solutions. Important parameters are the initial temperature $T_0$, the final temperature and the cooling scheme \citep{Kirkpatrick1983}. In the implementation, available on the EvoPER package, the default function for temperature update is $T = \alpha T$, being $\alpha$ the parameter controlling how fast the temperature is decremented. In addition, there are other methods readily available on the package for cooling and the users can also provide their own temperature decrement function. 

The Ant Colony Optimization algorithm is settled over the computational metaphor of the {\bf stigmergy} mechanism found in ant communities and used by the individuals for coordinating their activities in the search for food. Specifically, in the case of ant foraging behavior, the stigmergy is implemented by the pheromone reinforcing system where the most travelled way becomes the preferred one, owing to the proportional increment of pheromones deposited on the environment \citep{Dorigo1997}. The algorithm controls the convergence with the pheromone update and the pheromone evaporation processes. The evaporation avoids the rapid convergence to a local optimum \citep{Dorigo2006}. The standard version of ACO algorithm is well suited to discrete combinatorial problems but its application to continuous problems require some tweaking. Thus, to cope with these limitations an extension generalizing ACO for continuous domains and denominated $ACO_\mathbb{R}$ has been proposed \citep{Socha2008}. That extension, while keeping the underlying idea, replace the pheromone system by an equivalent structure called {\it solution archive} which stores the $s_l$ solutions, the results of $f(s_l)$ function  evaluations and finally the weight $\omega_l$ \citep{Socha2008}, for $i = 1, \dots, k$, where $k$ is an algorithm parameter for configuring the size solution archive. Finally, another component of algorithm is the Gaussian kernel which is sampled for update the solutions. The kernel contains $k$ Gaussian functions, one for each row $l$ in the solution archive.

The two metaheuristics introduced in this work, the {\it ees1} and {\it ees2} are simple strategies for tackling with the high computational cost of executing complex individual-based models the minimal number of iterations required for ensuring the algorithm has found an acceptable optimum value for the combination of model parameters when there is no information about what are their physically relevant ranges. We had developed the {\it ees1} and {\it ees2} for analyzing the parameter space of our own individual-based models \citep{PrestesGarcia2015a} \citep{PrestesGarcia2015b} of conjugation plasmid \citep{citeulike:13500899} dynamics within bacterial colonies. The underlying idea behind these metaheuristics is that we are interested on keeping the track and mapping the visited search space, rather than getting a point estimate for the best value, which may not be completely suitable for individual-based models because of the high stochasticity in the model output response. The obvious alternative for facing with the variability in the model output is increasing the number of replications for each parameter combination, but it would increase the execution time so much, rendering impractical the approach without parallelizing and distributing the load across several computer nodes. 

The {\it ees1} metaheuristic introduced in this work stands for {\bf EvoPER Evolutionary Strategy 1} and it is an instance of a custom evolutionary strategy which can be described by the commonly used notation as $(\mu + \rho_w \lambda)$-ES being $\mu = \lambda$, which basically means that every generation only the fittest individuals or, some suboptimal individuals selected with a probability $P$ taken from the parents and offspring collection, will become parents for the next generation. The mating selection process implemented in {\it ees1} choses the half of existing $\mu$ parents for being used in the recombination round. 

\begin{figure}[!hp]
\centering
\includegraphics[scale=0.7]{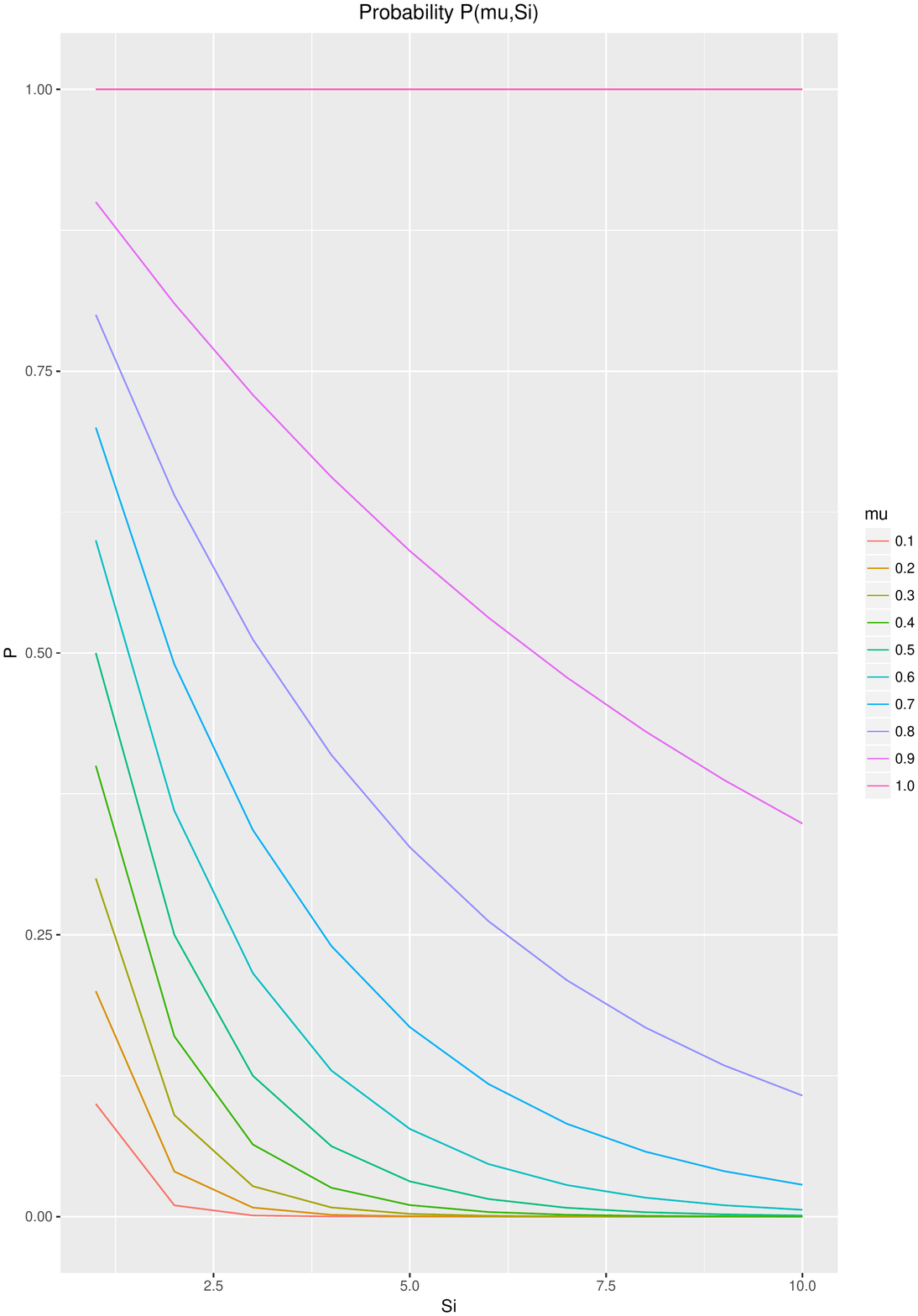}
\caption[Probability P(mu,Si)]{\label{fig:probability-mu} The probability $P(\mu)^k$ of selecting and individual having a suboptimal fitness value. }
\end{figure}

The algorithm has a parameter for parent selection which also uses the Greek letter  $mu$ but it must not be confounded with term used previously for describing the parent number in the evolutionary strategy descriptive notation. The parents are sorted by their fitness values and they are selected with an exponentially decreasing probability weight which is calculated using  the expression $P(\mu)^k, \forall k=1 \dots N$ where $ P(\mu)$ is the probability of selecting individuals with a suboptimal fitness, that is to say, when the values of  $ P(\mu)$ are small the solutions with the best fitness are string preferred and as the value $ P(\mu)$  tends to 1 the selection becomes a random process where the individuals are selected using an uniform deviate, see the Figure \ref{ fig:probability-mu} for visualizing how the parameter $\mu$ affects the probability of picking an individual having higher cost values.

The selected parents are used for calculating the geometric average which in turn serves as the centroid measure $\mathcal{C}$ for the recombination process in a sense that all individuals in the current solution population are recombined by calculating the arithmetic mean of the solution value and the $\mathcal{C}$. For executing the recombination process two different approaches are used, one of them is chosen with a probability $P=1/5$ and other with $1-P$. The first approach uses a {\it weight} value calculated as $w_k = f(x_k) / \sum_{k=1}^{N} f(x_k)$ for calculating the recombination component $\mathcal{R} =  ( x_k + \mathcal{C}) * w$. The new value of $x_k$ is the average $(x_k + \mathcal{R})/2$. The second approach selected with probability $1-P$ is simply the average of $x_k$ and $\mathcal{C}$. More details on the algorithm can be seen in Figure \ref{fig:EES-1}.

\begin{figure}[h]
\begin{lstlisting}[mathescape, language=Pascal]
$ N \leftarrow 10 $
$ \mu \leftarrow 0.3 $
$ \rho \leftarrow 0.01 $
$ \kappa \leftarrow 0.2 $
$ iterations \leftarrow 50 $

$ P \leftarrow $ LatinHypercubeSampling($\vec{p}, N$)
$ f \leftarrow $ Evaluate($P_0$)
$ S \leftarrow $ Sort($\{P_0, f\}$) 

while $i < iterations$
   # Mating selection
   $ mates \leftarrow $ select N/2 $S_i$,if($U < P(\mu,i)$) 
   
   $ r \leftarrow 1/(N/2) $
   $ G \leftarrow (\prod mates)^{r} $
   
   # Recombination
   for $i = 1 \dots N$
     $ w \leftarrow f(S_i) / \sum{S} $
     if( $ U < 1/5 $ )
        $S^{\prime}_i \leftarrow (S_i + S_i + G * w)/2 $ 
     else
        $S^{\prime}_i \leftarrow (S_i + G)/2 $ 			 
     end
   end   
   
   # Mutation
   $ S^{\prime} \leftarrow $ for each $(S^{\prime}_i)^p$ if( $U < \rho$ ) $ (S^{\prime}_i)^p + U $ 
   
   # Selection
   if $f(S^{\prime})$ beter than $f(S)$ or $U < \kappa$
      $ S \leftarrow S^{\prime} $
   end
end	
\end{lstlisting}
\caption[Evoper Evolutionary Strategy-1]{\label{fig:EES-1} . The pseudocode of evolutionary strategy 1 {\it ees1} .  The algorithm encompasses the standard components of an evolutionary strategy. First, the initial set of solutions is instantiated and evaluated. Subsequently, inside the main loop N/2 individuals are selected with a probability P, for estimating the geometric mean which will be employed for being recombined with the current solution population using a fitness weighted arithmetic mean.  The next step is the mutation process, consisting in making random changes in solution components with a probability $\rho$ from $1 to p$ variables, being $p$ the number of model parameters. Finally, if new solution improves the fitness, the current best solution is updated. The current best population solution $S_i$, for $i > 1$, is also updated with worst solutions with a probability $\kappa$. }
\end{figure}

The last component of {\it ees1} metaheuristics is the environmental selection where the fittest individuals are selected for being part of next generation replacing the current solution elements.  The best of all individuals from current population is always chosen for participating as a parent in the subsequent algorithm iteration and the remaining population components of suboptimal individuals can be replaced by the new individuals with a worst fitness value with a probability $P(S)$ which represents the strength of selective pressure. Thus, the range of $\kappa$ parameter may vary between $0 \le \kappa \le 1$, being the lower bound the absence of selective preference and the upper bound the maximum selective pressure over the population. In other words, a value of $\kappa = 0$ means all elements but the best of current population are replaced by the offspring independently of their fitness. On the other hand, a value of $\kappa = 1$ implies that only individuals improving the fitness of current solution are picked up for become part of next generation. For other values in the $\kappa$ parameter range, the selection process execute downhill movements with a probability $P(1-\kappa)$.

The second algorithm {\it ees2} is a very simple metaheuristic which is based on gradually reducing the initial dimensions of search space. The algorithm works by taking samples points of parameter space using Latin hypercube sampling scheme and allowing them vary over their full range. Thus, for each iteration step, the initial set of problem solution points are further refined using the defined fitness metric as guidance for reducing the full span of variation of each of the analyzed input factors. The parameters of this algorithm are the population size $N$, and the fraction of population size $\rho$ which will be used for calculating the new ranges for model parameters. The default values of these parameters are $N=100$ and $\rho=0.05$ but it can be tweaked depending on the number of input parameters being estimated.

\begin{figure}[h]
\begin{lstlisting}[mathescape, language=Pascal]
$ \rho \leftarrow 0.25 $
$ N \leftarrow 20 $
$ iterations \leftarrow 30 $
$ k \leftarrow trunc(N*\rho) $
$ r \leftarrow 0.5 $

$ P \leftarrow $ LatinHypercubeSampling($\vec{p}, N$)
$ f \leftarrow $ Evaluate($P$)
$ S \leftarrow $ Sort($\{P, f\}$) 

while $i < iterations$
	$ P	\leftarrow S(\{P, f\})$
	$ \vec{m_1} \leftarrow (min(P) \times max(P))^r $
	$ \vec{m_2} \leftarrow 1/2 |min(P) + max(P)| $
	
	if $U < 1/5$
		$ \vec{p}_{min} \leftarrow \vec{m_1} - \vec{m_2} - U $
		$ \vec{p}_{max} \leftarrow \vec{m_1} + \vec{m_2} + U $
	else
		$ \vec{p}_{min} \leftarrow \vec{m_1} - \vec{m_2} * U_{1 \dots 3} $
		$ \vec{p}_{max} \leftarrow \vec{m_1} + \vec{m_2} * U_{1 \dots 3} $
	end

	$ P \leftarrow $ LatinHypercubeSampling($\vec{p}, N$)
	$ f \leftarrow $ Evaluate($P$)
	
	$ S \leftarrow S $ add $\{P, f\}$
	$ S \leftarrow $ Sort($S$) 
	$ i \leftarrow i + 1 $
end	
\end{lstlisting}
\caption[Evoper Evolutionary Strategy-2]{\label{fig:EES-2} The complete pseudocode for {\it ees2} metaheuristic. This algorithm uses the Latin hypercube sampling together with a solution cost metric for reducing the solution search space towards the possible solution zone. }
\end{figure}

The Figure \ref{fig:EES-2} shows the general outline for the algorithm. As can be seen, the initialization section generates a sampling of problem space using Latin hypercube using the initial vector of input parameters $\vec{p}$ and the population size $N$. The input parameter vector $\vec{p}$ contains tuples with the range of values for input factors. Once initialized, the fitness function is evaluated and the results are added to solution $S$ and sorted using as sort key the fitness value. Thus, the best input parameter combinations for the problem are on the initial rows of $S$. The solution $S$ is a matrix with $m = N$ rows and $n = |\mathbb{X}| + 1$, being $N$ the parameter population size and $\mathbb{X}$ the set of input parameters, hence every row of $S$ have an instance problem input parameters and the fitness value for that parameter combination. The next section is the algorithm main loop where the search region is updated every iteration considering the first $k$ values from solution matrix. The number $k$ is calculated using the parameter $\rho$ and usually should be between 5-10\% of population size $N$. With the subset of solution matrix, the value of interval $I$ is calculated as the arithmetic average for the minimum maximum values of first $k$ elements of solution matrix $S$.  The minimum and maximum values of parameter range is then calculated using the average of first $k$ values of solution matrix $S$, the interval and a small random perturbation, then the new vector of input parameters $\vec{p}$ is used for generating the new round of sampling points using the Latin hypercube. The new population is evaluated and combined with the current solution $S$. The new solution is sorted and new iteration takes place using again the first $k$ values from solution matrix. As can be observed the algorithm is pretty simple but effective for mapping promising zones of solution space with a relative few number of iterations. It has not been extensively tested yet but when applied to standard optimization problems produce consistent results.   

The package was designed following an object-oriented approach, being structured around an entry point function and a class hierarchy representing an abstraction for the objective function to be optimized and by class encapsulating return type for the optimization methods. The classes abstracting the objective function are the basic input for the optimization algorithms available on the EvoPER package and can be extended for supporting other simulation platforms.  

There is a parent class called {\bf ObjectiveFunction} with two subclasses, namely the {\bf PlainFunction} and the {\bf RepastFunction}. The purpose of the first subclass is to allow the user to run the optimization algorithms using their own mathematical functions which can be useful for testing purposes or for wrapping other types of simulation subsystems. The second subclass encapsulates the Repast model calls for executing the chosen optimization algorithm for estimating the model parameters. The entry point function returns an object instance of {\bf Estimates} class. A brief description of package classes and the main methods is given in Table \ref{tab:classes} but for a complete and updated reference please refer to the package manual.

\begin{center}
\fboxsep0pt%
\fcolorbox{blue!30}{blue!30}{%
\parbox{\textwidth}{%
\fboxsep5pt
\centering
\captionof{table}{\label{tab:classes} The partial structure of EvoPER classes for encapsulating the target of parameter estimation.}
\begin{tabular}{llp{8.0cm}}
\toprule
Class name & Methods & Description \\ 
\midrule
ObjectiveFunction 	&						& The base class in hierarchy providing the skeleton for running the optimization algorithms.					\\ 
 					& Parameter				& The objective of this method is adding parameter with range between a minimum and a maximum value. 
 											  	The parameters must coincide with those defined in the model.												\\ 
 					& getParameter			& Returns a previously defined parameter. The companion method {\bf getParameterNames} returns a list with the 
 												names of all user defined parameters.																		\\ 
					& Evaluate				& The {\bf Evaluate} method is responsible for wrapping the objective function evaluation and must be overridden
											  	by all those classes extending the parent class {\bf \mbox{ObjectiveFunction}}.								\\  					 					
					& EvalFitness			& Execute the objective function and returns the results. The user code must prefer this method for executing 
												the objective function.																						\\ 
 					& RawData				& Return the complete raw output of objective function whenever it is available.								\\
					& stats					& Provide some basic statistics for the {\bf ObjectiveFunction} execution.										\\
PlainFunction 		&   					& Allows the optimization of plain functions  implemented in R. 												\\ 
					&  initialize			& The initialize method must be overridden in subclasses of {\bf ObjectiveFunction} and it is responsible for 
												bootstrapping the real implementation of target function. In the case of {\bf PlainFunction} it requires 
												any user provided R function as parameter.  																\\
					&  Evaluate				& Override superclass method with the specific function call.														\\  					 					
RepastFunction 		&  						& Wrapper for the Repast Model objective function.																\\ 
					&  initialize			& This method is a wrapper initializing the Repast Model constructor. Requires the model directory, an 
												aggregated data source, the total simulation time and a user defined cost function.							\\
					&  Evaluate				& Override superclass method with the specific function call.													\\  					 					
Estimates			& 						& The {\bf Estimates} class serves as the standardized return type for all optimization methods available on 
												the	package.  The initialized instance of this class stores the values of best value ever found during 
												the metaheuristic execution, the list of best values for every algorithm iteration and finally the 
												complete collection of all points which have been visited in the solution space which can be particularly 
												useful for mapping the promising zones solution space.														\\ 		
					& getBest				& This method returns the best value ever found for the objective function.										\\
					& getIterationBest		& Returns a list with the best values found for every iteration.												\\
					& getVisitedSpace		& The method returns a collection which contains the results of all evaluations of the objective function.		\\
\bottomrule
\end{tabular}
}%
}
\end{center}

The object-oriented approach allows the easy extension of the package for other types of Individual-based modeling tools or methods. As can be seen in Table \ref{tab:classes} the only requirement to apply the methods contained in the EvoPER package is to extend the {\it ObjectiveFunction} class and override the Evaluate method to support the new parameter estimation target. One of the useful aspects of EvoPER implementation is the possibility to specify constraints in the search space by individually setting lower and upper bounds for every parameter being analyzed using the  {\it ObjectiveFunction\$Parameter(name, min, max)} method. That is an important point for limiting the parameter values only to the acceptable biological range.  

The workflow for carry out the parameter estimation consists in a simple sequence of steps. First, an object instance of any {\it ObjectiveFunction} subclasses must be created and properly initialized. As mentioned previously, currently we have two options available for parameter estimation: one for simple functions which could be used for testing purposes ({\it PlainFunction}) and another for estimating parameters of Repast models ({\it RepastFunction}).  Once the objective function has been initialized, the required parameters must be provided with the appropriate lower and upper bounds. Finally, the {\it extremize} function can be applied to the previously defined function. The required parameters are the optimization method and the objective function instance. The function has a third optional parameter for providing the custom options for the underlying optimization method. 

The optimization functions available are shown in the Table \ref{tab:functions} for providing an overview on the package contents. The EvoPER package is still in and earl phase of development therefore the list could change over the time. The package manual will be the most updated source of information for the package contents.

\begin{center}
\fboxsep0pt%
\fcolorbox{blue!30}{blue!30}{%
\parbox{\textwidth}{%
\fboxsep5pt
\centering
\captionof{table}{\label{tab:functions} The overview of the most relevant EvoPER functions. The current a list with implementation of metaheuristic methods for parameter estimation.}

\begin{tabular}{lp{11.4cm}}
\toprule
Function 				& Description 	\\ 
\midrule
extremize				& This is the entry point function for all available parameter estimation methods and should be preferred instead of direct call to the underlying functions. It has three parameters, being the first two of them required and the third optional. The first parameter is a string indicating the metaheuristic algorithm, currently accepted values are $("pso"|"saa"|"acor"|"ees1"|"ees2")$ for particle swarm, simulated annealing, ant colony respectively, the evolutionary strategy 1 and evolutionary strategy 2. The next version will include also genetic algorithms (GA) and Tabu Search (TS). The second is an instance of the objective function and finally, the third one is an instance of a subclass if {\bf Options} class specific for algorithm. If not provided the default options for the metaheuristic will be used. The extremize returns an initialized object instance of {\bf Estimates} class containing the results for the optimization method.																															\\

abm.acor				& The abm.acor implements the {\bf Ant Colony Optimization} for continuous domains.  The function requires an instance of a subclass of {\bf ObjectiveFunction} and an optional parameter with an instance of {\bf OptionsACOR}. Currently there are two subclasses of ObjectiveFunction, one for optimizing plain R functions ({\bf PlainFunction}) and another for Repast Models ({\bf RepastFunction}).												\\

abm.pso					& The function call for running the {\bf Particle Swarm Optimization} method. It is necessary to provide a subclass of the {\it ObjectiveFunction} and optionally an instance of {\bf OptionsACOR}. If not options are given a default instance will be used for the maximum iterations, the swarm size, the acceleration coefficients, the inertia weight or constriction coefficient and finally the neighborhood type.							\\

abm.saa 				& This is the implementation of {\bf Simulated Annealing} algorithm and identically as in the previous cases the function requires an instance of the objective function and accepts an instance of {\bf OptionsSAA}. The options class have acceptable the default values for the initial temperature, the minimum temperature, the temperature length, the cooling ratio, the neighborhood distance as a fraction of parameter range and the neighborhood function. 																																	\\

abm.ees1				&The {\bf EvoPER Evolutionary Strategy 1 (ees1)} is a simple evolutionary strategy which uses the geometric mean as the focal point for constructing the recombination model for the next generation of candidate solutions. The metaheuristic allows the configuration of several parameters, namely the solution size ($N$), the mating selection strength ($\mu$), the mutation rate $rho$ and the selective pressure ($\kappa$). The default values can be changed by providing an instance of options class {\bf OptionsEES1} which the desired values.																													\\

abm.ees2				& That is not exactly an evolutionary strategy {\it stricto sensu} because the new generation solution is not created directly from parent solution but instead, parents are used for searching the new range of solution parameter space. The algorithm is based on reducing the initial parameter space and generating new solutions with the new ranges for each iteration. The solutions are generated using the {\it Latin hypercube sampling} scheme. The configurable parameters are basically the population size $N$, the number of algorithm iterations and the selection ratio $\rho$ which allows the specification of a fraction of $N$ for estimates the new boundaries. The default values for both parameters are 10 and 100 respectively. For modifying these settings, it is necessary to provide an instance of options class {\bf OptionsEES2} which the desired values.  It is intended to provide an acceptable approximate for parameter estimation in fewer model executions.  												\\

\bottomrule
\end{tabular}
}%
}
\end{center}

The {\bf particle swarm optimization} metaheuristic implementation requires a topological neighborhood function which provide the structure for the swarm particles allowing the algorithm to select the best position in the solution search space. The package provides three different implementations for the neighborhood selection: \mbox{\bf pso.neighborhood.K2}, \mbox{\bf pso.neighborhood.K4} and \mbox{\bf pso.neighborhood.KN}. The first topology function returns two neighbors of solution particle $x_i$, where the neighbors are the particles $x_{i-1}$ and $x_{i+1}$ using a ring topology \cite{Zambrano-Bigiarini2013}. The second function returns four neighbors of particle $x_i$ using a von Neumann neighborhood function. Finally, the last function returns a complete graph with the whole set of particles. The default implementation uses the entire population as the neighborhood. In addition to these functions, it is also possible to provide a user defined neighborhood function creating a non-default instance of the {\bf OptionsPSO} class and passing the reference to the alternative implementation using the method {\bf neighborhoodFunction()} of Options class. The neighborhood function signature is shown in Figure \ref{fig:pso-neighborhood}. The functions is invoked inside the PSO code and which pass the position of current particle $i$ the size of particle population $n$; the function must return a collection of integers with the neighbor positions, grouped with the R $c()$ call.  

\begin{figure}[h]
\begin{lstlisting}[basicstyle=\scriptsize, language=R]
myneighborhood<- function(i,n) {
	c()	
}
\end{lstlisting}
\caption[PSO neighborhood]{\label{fig:pso-neighborhood} The function signature for the custom particle swarm neighborhood. The function can return any number of neighbors from $i$ to $n$ and the returned values must not be greater than $n$. }
\end{figure}

Most of the aspects implemented in the optimization code are standard and, perhaps the only points which are specific to the EvoPER package, are the neighborhood function for {\it pso.neighborhood.K4} and {\it saa.neighborhood}. The von Neumann neighborhood for particle swarm optimization is generated using a topology created converting the linear collections of particles to a matrix using the R code \verb| m <- matrix(seq(1,N),nrow=(ceiling(sqrt(N))))| where $N$ is the swarm size.

The case of generating the neighborhood solutions for simulated annealing has been attacked using the following logic for generating new solutions: first pick randomly the parameters to be perturbed\footnote{The neighborhood functions currently implemented allows choosing from 1, 1/2 n or n, being n the number of parameters which will be perturbed. The implementation can be easily extended for accommodating any user defined neighborhood algorithm.} and update them using two different paths selected randomly, preferring the second with a probability $P = 0.8$. The first of then is based on drawing a number from a normal deviates given by the expression $S' = \Delta Z + \bar{S}$ where $S'$ represents the new solution, $\Delta$ the standard deviation calculated as the range of parameter $p_k$ multiplied by the algorithm parameter distance $d=0.5$ (default value) finally, the $\bar{S}$ is the arithmetic average between the minimum and maximum  allowed values of parameter $p_k$.  The second one uses the expression $S' = S + S * U(-1,1)$ where $S'$, $S$, $U$ are respectively the new neighbor solution, the current solution, a uniform random number between $[-1, 1]$. 

The {\bf simulated annealing} algorithm also uses needs a function for generating other points in the solution space close to the actual current solution. The currently available neighborhood functions for perturbing the best solution are: \mbox{\bf saa.neighborhood1}, \mbox{\bf saa.neighborhoodH} and \mbox{\bf saa.neighborhoodN}. The difference between these implementations is basically the number of problem dimensions to be perturbed. Thus, the first function alters just one element of current solution, the second changes the half of solution elements at a time and finally the last one modify all dimensions of a solution. The solution components to be perturbed are chosen randomly. Again, it is possible for the package users provide their own implementation for the neighborhood function. 

The package provides acceptable default values for most of parameters related to the optimization method in use. In spite of the fact that the parameter estimation functions can be called directly, the users should use the function {\it extremize(m, f, o)} which is the standard entry point for the optimization methods. As has been mentioned previously, the function has three parameters, which are respectively the method ({\it m}), the objective function ({\it f}) and the options ({\it o}). Only the first two are required and the third is optional. When the options parameter is not provided the default values are used. If setting different from the default values are required, the user must pass an instance of the corresponding option class. For example, if more iterations are required for PSO method an instance of {\it OptionsPSO} must be created and the method {\it setValue("iterations", value)} with the appropriate value.  Many other parameters can be customized in order to fit the specific needs for the model being analyzed such as the neighborhood functions or the temperature update for the simulated annealing.

\section{Discussion}

In this section, we will show some small and illustrative examples about how to use the EvoPER package for estimating the parameters of different kinds of models.  The first example includes the parameter estimation required minimizing the standard functions employed for testing optimization methods. The second example shows show to adjusting the output of a real individual-based model to match the experimental data. The next one is on how to tune the model output for oscillating with any specific user defined period. Finally, the last one give an example on how to explore the parameter solution space for getting a landscape with suitable solutions for the problem being addressed.

\subsection{Optimizing simple functions}

It is worth mentioning that although the package is oriented to the application of evolutionary optimization methods to the parameter estimation of models developed using Repast Simphony it can also be used to minimize mathematical functions as well as, extended for other individual-based modeling frameworks. In the following example shown in Figure \ref{fig:example1} we demonstrate the package usage applying it to the two variables Rosenbrock's function.

\begin{figure}[h]
\begin{lstlisting}[basicstyle=\scriptsize, language=R]
# Step 0
rm(list=ls())
library(evoper)
set.seed(161803398)

# Step 1
rosenbrock2<- function(x1, x2) { (1 - x1)^2 + 100 * (x2 - x1^2)^2 }

# Step 2
objective<- PlainFunction$new(rosenbrock2)

# Step 3
objective$Parameter(name="x1",min=-100,max=100)
objective$Parameter(name="x2",min=-100,max=100)

# Step 4
results<- extremize("pso", objective)

\end{lstlisting}
\caption[Rosenbrock ]{\label{fig:example1} A simple example for minimizing the Rosenbrock's function using the EvoPER package.}
\end{figure}

As can be seen in Figure \ref{fig:example1} the {\bf step 1} shows the definition of a simple function to be minimized; the {\bf step 2} demonstrate how to create an instance of {\it PlainFunction} class; in the {\bf step 3} the parameter ranges for each function's parameter is provided and finally in the {\bf step 4} the EvoPER {\it extremize} function is used to minimize the objective function.  The results of running the example are shown in Figure \ref{fig:example11} where can be seen the estimated parameters, the value of fitness function, the execution time and the number of times the function has been evaluated.

\begin{figure}[h]
\begin{lstlisting}[basicstyle=\scriptsize, language=R]
> system.time(results<- extremize("pso", objective))
   user  system elapsed 
   4.96    0.01    4.97 
> results$getBest()
        x1       x2 pset      fitness
1 1.002273 1.004707    9 7.561967e-06
> objective$stats()
     total_evals converged    tolerance
[1,]        2416         1 2.013409e-05
\end{lstlisting}
\caption[Running Rosenbrock]{\label{fig:example11} The R console output session showing the results of running the previous example. }
\end{figure}

\subsection{Tuning oscillations}

The oscillatory behavior is a structural component of many types of systems requiring timers for controlling and coordinating its processes. It is an integral part of several types of systems, ranging from electronic to ecological or biological processes which normally relies on circadian clocks for regulating faithfully all their internal activities and interactions with environment. Therefore, the design of synthetic biological oscillatory circuits is an important research subject and tuning these circuits for oscillating with a precise period a cumbersome and trial and error activity \cite{Khalil2010}.  Fortunately, tuning the model parameters for finding the desired oscillatory behavior can be expressed as am optimization problem which can be solved using evolutionary algorithms. It is worth to mention that the parameters estimated for any model are just starting point for the wet-lab work because the reality is extremely stubborn insisting in not working in line with the values estimated by the model. 

For illustrating how to turn that problem into an optimization problem we introduce an example shown in Figure \ref{fig:example3} where the cost function is crafted for tuning the model parameters in order to accomplish a specific output. Specifically, it is a simple toy model representing the Lotka-Volterra ordinary differential equation system, also known predator-prey is presented and we want to estimate the parameters required to make the output oscillate with a specific period. The model, despite of being developed for modeling the predator and prey relationship, has a broad range of applications and can be used for representing many types of ecological and biological interactions \cite{Shonkwiler:2008}. Additionally, the standard model can be extended for supporting $N$ species interactions.

\begin{center}
\fboxsep0pt%
\fcolorbox{blue!30}{blue!30}{%
\parbox{\textwidth}{%
\fboxsep5pt
\centering
\captionof{table}{\label{tab:period-tuning} The results of applying particle swarm optimization metaheuristic for period tuning. The table shows the optimum parameter values for the oscillation period.}
\begin{tabular}{p{3cm}llllp{1cm}r}
\toprule
Period 	& c1 		& c2 		& c3 		& c4 		&	& cost 						\\ 
\midrule
12 		& 1.798102	& 1.618035 	& 1.192361 	& 1.453045 	& 	& 0 						\\
24		& 0.675586 	& 1.375913 	& 1.169076 	& 0.8311187 &  	& 0.04166667				\\
48		& 0.4558475 & 0.4602389 & 1.192546  & 0.5483637	& 	& 0							\\
72		& 0.3297914 & 0.4675479 & 1.650108  & 0.778639 	& 	& 0							\\
\bottomrule
\end{tabular}
}%
}
\end{center}

\begin{figure}[!hp]
\centering
\includegraphics[scale=0.7]{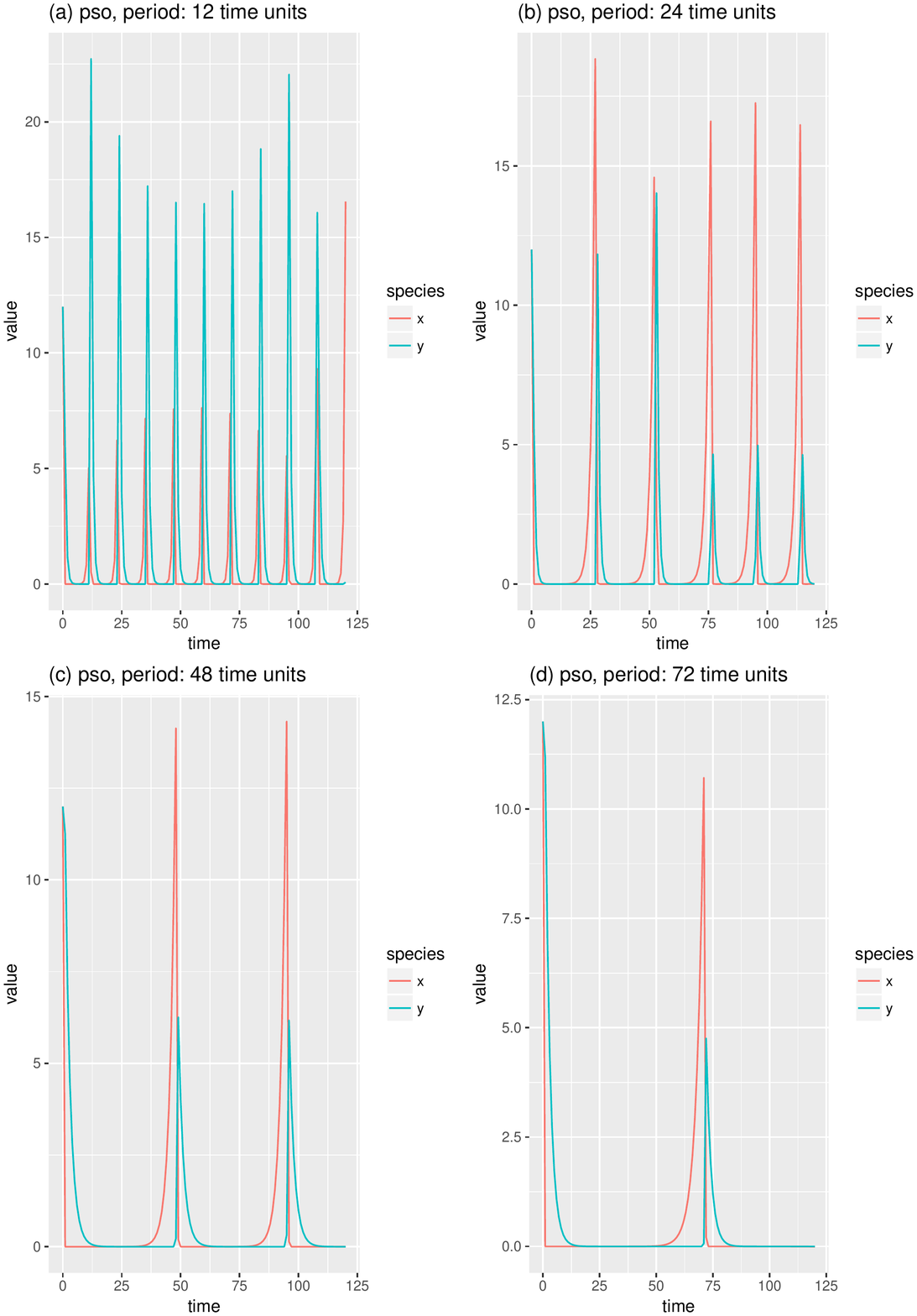}
\caption[Predator/Prey period tuning]{\label{fig:period-tuning} An example of tuning the oscillation periods of predator-prey model. In this figure, we can observe how {\it x} and {\it y} species, respectively the prey and predator components oscillates with different periods. The example uses the particle swarm optimization metaheuristic for finding the parameter combination required for making the model produce oscillations with periods of 12, 24, 48 and 72 time units as can be seen respectively in subfigures (a), (b), (c) and (d). As can be observed the objective function of metaheuristic can be tweaked for generating any desired output behavior. The most common one is to assess the quality of fit between simulated and experimental data but it is no limited and can be used to find parameter combinations which generate practically any global behavior.}
\end{figure}

The standard predator-prey model has four parameters which are necessary to estimate as can be seen in Equation \eqref{eqn:predator-prey},

\begin{align}
\label{eqn:predator-prey}
\frac{dx}{dt} &= c_1 x - c_3 x y \\
\frac{dy}{dt} &= -c_2 y + c_4 x y \nonumber. 
\end{align}

where the terms $c1$, $c2$, $c3$ and $c4$ which represent respectively the growth rate of prey, the predation rate, the predation effect on predator growth rate and finally the death rate of predator. The session output is of a model execution for tuning an oscillation period of  72 time units is presented in Figure \ref{fig:example31} where the values for the parameters required to produce oscillations with the desired period are shown. Additionally, the Table \ref{tab:period-tuning} shows the complete results for all periods for which the model parameters have been estimated. The Figure \ref{fig:period-tuning} shows graphically the results for the tuning the model parameters for producing an oscillatory behavior with approximate periods of 12, 24, 48 and 72 time units.

\begin{figure}[h]
\begin{lstlisting}[escapeinside={(*}{*)}, basicstyle=\scriptsize, language=R]
# Step 0
rm(list=ls())
library(evoper)
set.seed(2718282)

# Step 1
(* \dots *)
f0.periodtuningpp<- function(x1, x2, x3, x4, period) {
  v<- predatorprey(x1, x2, x3, x4)
  np<- naiveperiod(ifelse(v[,"y"] < 0, .Machine$double.xmax, v[,"y"]))
  rrepast::AoE.NRMSD(np$period, period)
}
(* \dots *)

# Step 2
objective<-PlainFunction$new(f0.periodtuningpp72)

# Step 3
objective$Parameter(name="x1",min=0.2,max=2)
objective$Parameter(name="x2",min=0.2,max=2)
objective$Parameter(name="x3",min=0.2,max=2)
objective$Parameter(name="x4",min=0.2,max=2)

# Step 4
results<- extremize("pso", objective)
\end{lstlisting}
\caption[Predator Prey]{\label{fig:example3} Tuning the oscillation period of predator-prey model. The listing has five sections identified by the tags {\it Step 0} to  {\it Step 4}. The first section consists in loading the library and setting the random seed. The next section is where the cost function is defined, consisting in solving the initial value problem with provided parameters and using the results of ODE for feeding the function named {\bf naiveperiod} for finding the periods in the differential equation output which is later compared with the reference period using a normalized root mean square deviation (AoE.NRMSD). The subsequent sections encompasses: the initialization of function to be optimized which is a wrapper for the previously defined cost function; the definition of range of variation for the model parameters and finally the application of the metaheuristic with the {\bf extremized} function call.}
\end{figure}

\begin{figure}[h]
\begin{lstlisting}[basicstyle=\scriptsize, language=R]
> system.time(results<- extremize("pso", f))
   user  system elapsed 
 225.89    0.01  226.16 
> results$getBest()
         x1        x2       x3       x4 pset fitness
1 0.3297914 0.4675479 1.650108 0.778639    1       0
> f$stats()
     total_evals converged    tolerance
[1,]         672         1 2.013409e-05
\end{lstlisting}
\caption[Running Rosenbrock]{\label{fig:example31} The R console output session showing the results of running predator-prey model in Figure \ref{fig:example3}. }
\end{figure}

\subsection{Exploring the solution space}

The particularities of individual-based models which make them so appealing for modeling populations and ecosystems, such as the structural realism, the predictive power, the individual level stochasticity and the emergence of complex global dynamics from the elemental interactions \citep{Grimm2016} also implies that that parameter estimation become a complicated matter even when using approximated techniques as those presented in this work. Even simple models contain many levels of uncertainty and certainly presenting nonlinear behavior and possibly discontinuities consequently it is very complicated, using a computationally tractable number of model executions, to make sure that the solution converges successfully to an optimum which is close to the better solution. Normally, modelers have not enough information about all model parameters and it is no uncommon to make assumptions or educated guesses for the acceptable ranges considering the physical or biological constraints. Off course, it is not random choice but it is far from being a perfect process and, despite of guessed parameter ranges are hopefully within the same order of magnitude of their real values, usually they may diverge by a factor of two three \citep{ Milo16} which may produce odd results when adjusting several parameters.

It is worth noting that it should not be expected a perfect match between the model predictions and the experimental data, consequently it is very unlikely that optimization algorithms converge using the tolerance levels used normally for numerical optimization of plain functions. One way to tackle with this situation is defining the objective function for the parameter estimation using a categorical approach with a not much strict range of acceptance but that may lead to loosing information which may be relevant and giving the false felling that the right parameter combination has been found.  That is serious issue which may render impossible to draw any conclusion from the parameter estimation results. A possible alternative is not relying exclusively on the best value ever found for the cost function but instead, leveraging the intermediate results for analyzing the problem using it for building a landscape of solution space. The metaheuristics described in this work have a slight modification for saving the partial best results for every interaction and the complete set of points visited in the problem solution space which are made available as two methods of {\it Estimates}  class, respectively {\it getIterationBest()} and {\it getVisitedSpace()}. Using these two methods the solution space can be mapped for viewing the most promising zones. The code in Figure \ref{fig:solution-space1} shows how to generate a contour plot for the solutions space for a four variables instance of {\it Rosenbrock} function.

\begin{figure}[h]
\begin{lstlisting}[basicstyle=\scriptsize, language=R]
# Step 0
rm(list=ls())
library(evoper)
set.seed(161803398)

# Step 1
objective<- PlainFunction$new(f0.rosenbrock4)

# Step 2
objective$Parameter(name="x1",min=-100,max=100)
objective$Parameter(name="x2",min=-100,max=100)
objective$Parameter(name="x3",min=-100,max=100)
objective$Parameter(name="x4",min=-100,max=100)

# Step 3
results<- extremize("acor", objective)

# Step 4
p1<- contourplothelper(v$getVisitedSpace()[1:200,],"x1","x2","fitness")
p2<- contourplothelper(v$getVisitedSpace()[1:200,],"x2","x3","fitness")
p3<- contourplothelper(v$getVisitedSpace()[1:200,],"x3","x4","fitness")
p4<- contourplothelper(v$getVisitedSpace()[1:200,],"x4","x1","fitness")

\end{lstlisting}
\caption[Solution space of Rosenbrock 4]{\label{fig:solution-space1} Exploring the solution space for Rosenbrock function of four variables using the Ant Colony Optimization for continuous domains (acor) algorithm. The sequence of steps is practically the same presented previously with the exception of {\it Step 4} which show the generation of four contour plots using the first 200 values retuned by the {\it \mbox{getVisitedSpace()}} method which are sorted in ascendant order by the fitness value.}
\end{figure}

\begin{figure}[ht]
\centering
\includegraphics[scale=0.7]{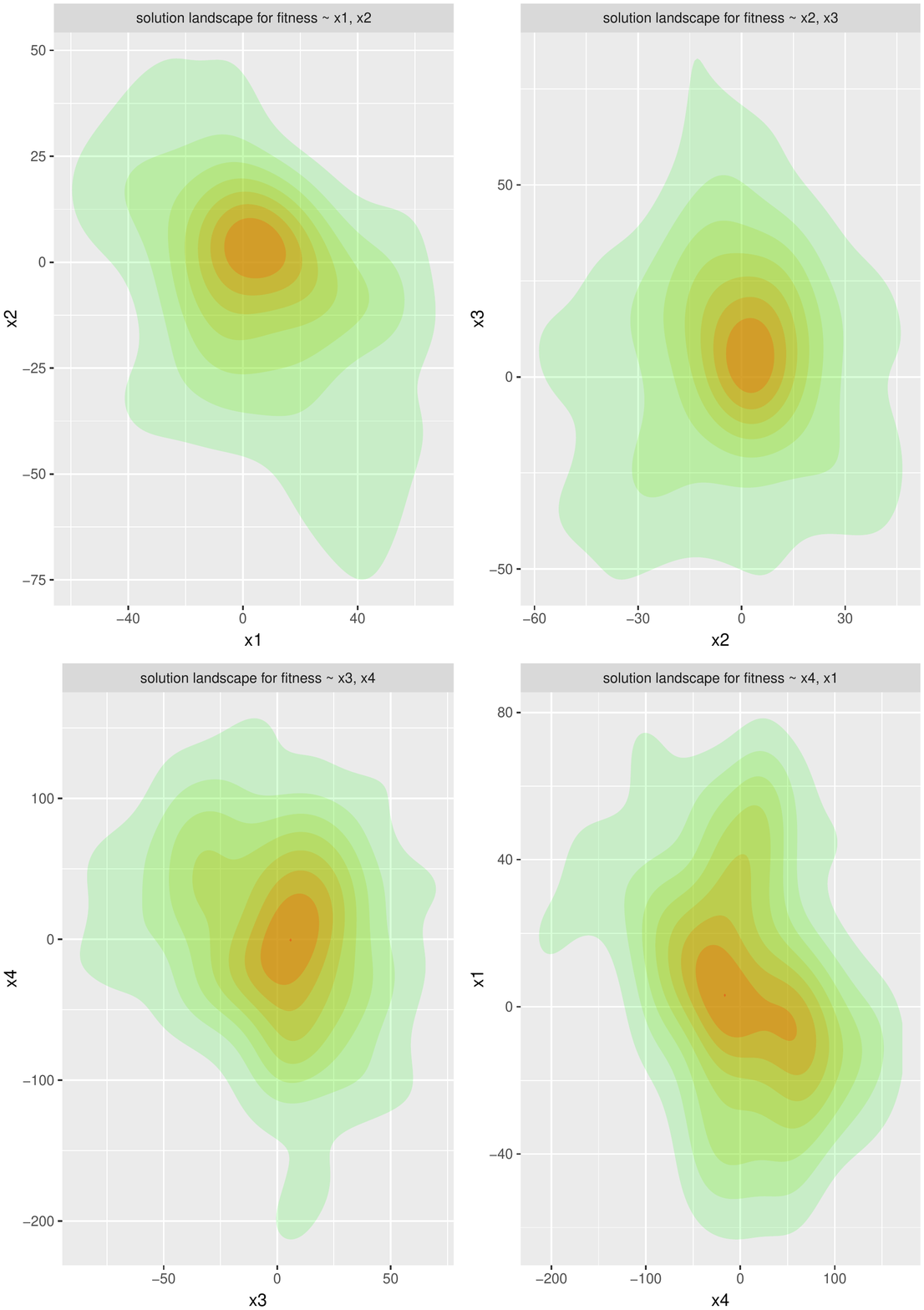}
\caption[Exploring ACOR solution landscape]{\label{fig:exploring-solution-1} Exploring solution landscape for visited space generated during the execution of {\it Ant Colony for Continuous domains (acor)} algorithm. The four contour plots provides a panoramic view for the fitness surface of the variable pairs (x1,x2), (x2,x3),  (x3,x4) and (x4,x1). The contour curves are employing a color scheme, from light green to red for indicating the cost value, which means respectively the worst and the best fitness for the function.}
\end{figure}

\begin{figure}[ht]
\centering
\includegraphics[scale=0.7]{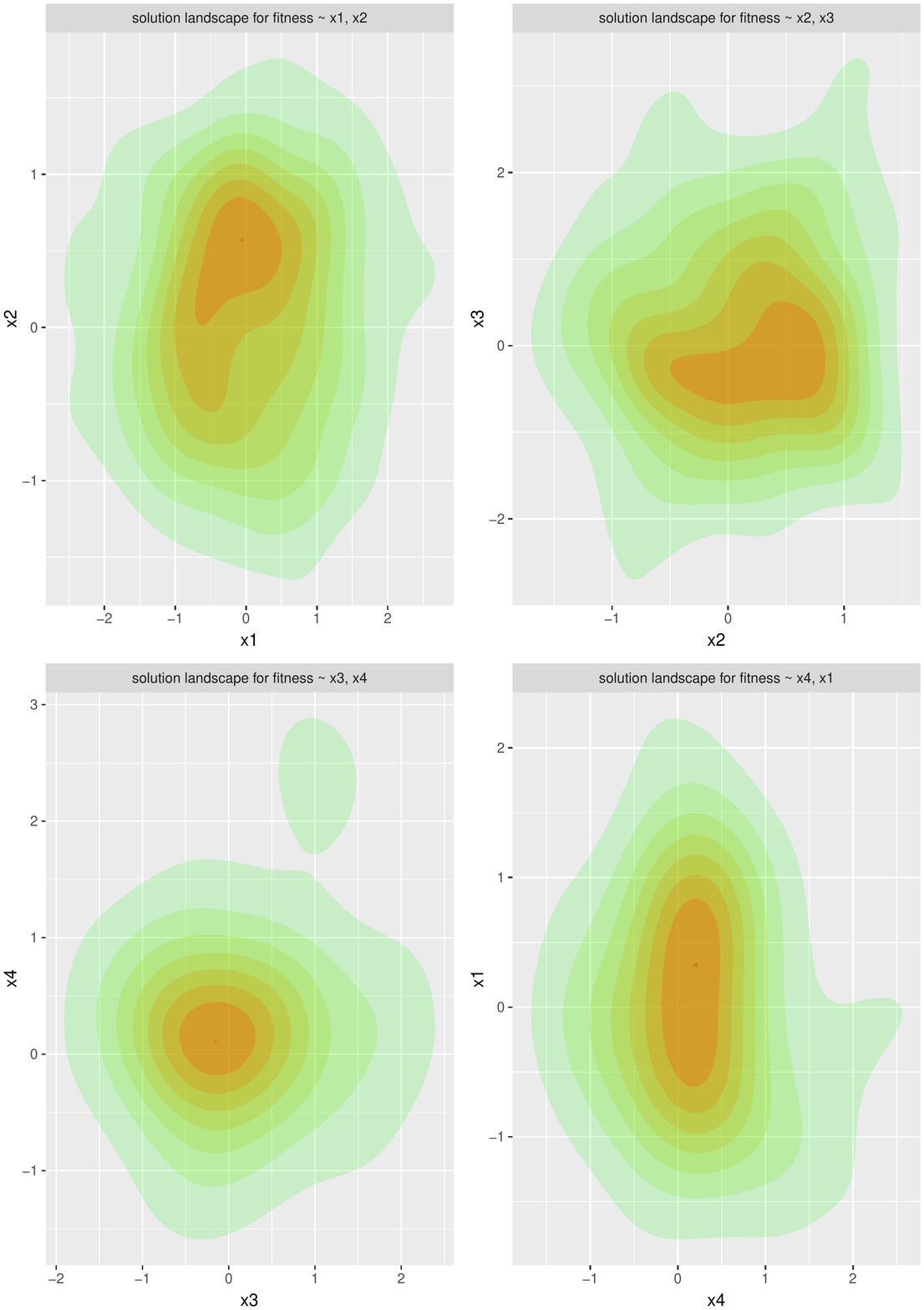}
\caption[Exploring EES2 solution landscape]{\label{fig:exploring-solution-2} Exploring solution landscape for visited space generated during the execution of {\it EvoPER Evolutionary strategy 2 (ees2)} algorithm. The four contour plots provides a panoramic view for the fitness surface of the variable pairs (x1,x2), (x2,x3),  (x3,x4) and (x4,x1). The contour curves are employing a color scheme, from light green to red for indicating the cost value, which means respectively the worst and the best fitness for the function. }
\end{figure}

The complete plot generated with the fourth step of command sequence provided in  \ref{fig:solution-space1} is shown in Figure \ref{fig:exploring-solution-1}. These contour plots facilitate mapping and visualizing the promising zones of solution space using the fitness of solution generated as the $z$ value. The model parameters are disposed two by two making easy to find the zone where the best solution of adjacent parameters may be possibly situated. The algorithm employed was the Ant Colony Optimization for continuous domains which have not converged and the complete execution has required approximately 32K model evaluations as the default options for the metaheuristic are 500 iterations using a population of 64 ants. The previous value is certainly not acceptable for a costly individual based model which may require from several hours to days for such high number of evaluations. Thus, it is necessary to tune the metaheuristics for reducing the total number of iterations or alternatively using the algorithm introduced in this work {\it ees2} for making the initial tour to the solution landscape. The {\it ees2} is not intended to find the minima but instead it is well suited for partitioning the problem space towards the good solution zones. The \ref{fig:exploring-solution-2} shows the contour plots generated using {\it ees2} which required just 600 model evaluations for reducing the solution zone. The global minimum for the Rosenbrock function of $N$ variables is zero and is found setting all variables to 1.

The Figures \ref{fig:exploring-solution-1} and \ref{fig:exploring-solution-2}  are demarcating the possible zones were the problem solution can be found. In the case of Rosenbrock test function, which has been used in these examples, the best solution is known a priori to be zero when all parameters are 1. Thus, looking on the first plot, can be easily observed that the hot zones marked with best fitness are those corresponding the best problem solution. Nonetheless, the complete solution landscape encompasses a very wide zone when compared to the solution estimated using the {\it ees2} which shows a much more restricted portion of solution space. It is important to note that real models normally do not generate results so evident as those generated with test functions. Usually, the real individual-based models will show discontinuities and possibly more than one zone with good fitness values owing to the nonlinear or second order interaction between model parameters.

\subsection{Comparing metaheuristics}

The task of choosing the most suitable metaheuristic for the parameter estimation problem is not an easy one, mainly because the available algorithms behave differently depending on the problem type. It is also a consequence of intrinsic stochasticity present on individual-based models as well as the random nature of algorithms itself.  Therefore, it is hard to provide general recipes for deciding what method is right for a particular problem instance. The algorithms enclosed in this work are tuned with those parameters adequate for general cases but some tweaking may be required for achieving the best results. Specifically, the optimization metaheuristics can be very sensitive to the neighborhood structure and to the parameters controlling the balance between local search which accelerate the convergence speed and breadth of search which may avoid to get stuck in local optima failing to converge to the best solution. It may be necessary some trial and error approach, testing different algorithms, observing the convergence speed, the value of cost function and then tuning the algorithm parameters accordingly.

The application of optimization metaheuristics to individual-based models, as mentioned previously, poses an additional problem because every model execution is computationally costly when compared to other models types and the cost of algorithm itself can be neglected. Therefore, the one of the most important factors for selecting an algorithm is the minimal number of evaluations of objective functions which are needed for finding an acceptable solution satisfying the optimization target. Bearing this in mind, this section provides a systematic comparison between some of metaheuristics mentioned in this work. The metaheuristics have been compared using the functions known as {\bf Cigar}, {\bf Schaffer}, {\bf Griewank} and {\bf Bohachevsky} \citep{Qu2016} \citep{Jamil2013} which are standard test functions commonly employed for benchmarking the optimization algorithms. The benchmarks were performed using the four variables version of test functions and the experiments were replicated seven times using randomly selected initial random seeds\footnote{The details and the code used for the benchmark are enclosed along the package sources which are available on \url{ https://github.com/antonio-pgarcia/evoper }}. The summarized numerical results obtained from the benchmark process are shown on Table \ref{tab:benchmark1}.

\begin{center}
\fboxsep0pt%
\fcolorbox{blue!30}{blue!30}{%
\parbox{\textwidth}{%
\fboxsep5pt
\centering
\captionof{table}{\label{tab:benchmark1} The output of benchmarking metaheuristics algorithms. These results were produce using the function {\it compare.algorithms1} included in package distribution and are the average values of seven replications with different initial random seeds. The convergence values are the ratio of replications which actually converged over the total number of performed experiments. }
\begin{tabular}{p{3cm}lrrlp{1.5cm}r}
\toprule
Function 						& Algorithm 		& Evaluations 		& Convergence 		&  	&	& Fitness		\\ 
\midrule
Cigar							& saa				& 458.14 			& 1.0 				&	&	& 0.04702100 	\\
								& pso				& 2621.71  	 		& 1.0 				&	&	& 0.05033806 	\\
								& acor				& 2651.43   		& 1.0 				&	&	& 0.07268033 	\\
								& ees1				& 332.86    		& 1.0 				&	&	& 0.06308869 	\\
Schaffer						& saa  				& 649.57   			& 1.0 				&	&	& 0.07438586	\\
								& pso 				& 6269.71  			& 0.3 				&	&	& 0.57479139	\\
								& acor  			& 2249.14  			& 1.0 				&	&	& 0.08662474	\\
								& ees1  			& 495.71   			& 1.0 				&	&	& 0.08821531	\\
Griewank						& saa   			& 501.00   			& 1.0 				& 	&	& 0.04579390	\\
								& pso   			& 4214.86  			& 0.6 				& 	&	& 0.09860997    \\
								& acor  			& 3812.57  			& 1.0 				& 	&	& 0.07698225    \\
								& ees1  			& 308.57   			& 0.9 				&	&	& 0.07089164    \\
Bohachevsky						& saa   			& 258.14  			& 1.0 				&	&	& 0.05232314	\\
								& pso   			& 5053.71  			& 0.4 				&	&	& 0.44120018    \\
								& acor  			& 1362.29  			& 1.0 				&	&	& 0.04323902    \\
								& ees1  			& 258.57   			& 1.0 				&	&	& 0.06660127    \\
\bottomrule			
\end{tabular}
}%
}
\end{center}

\begin{figure}[ht]
\centering
\includegraphics[scale=0.7]{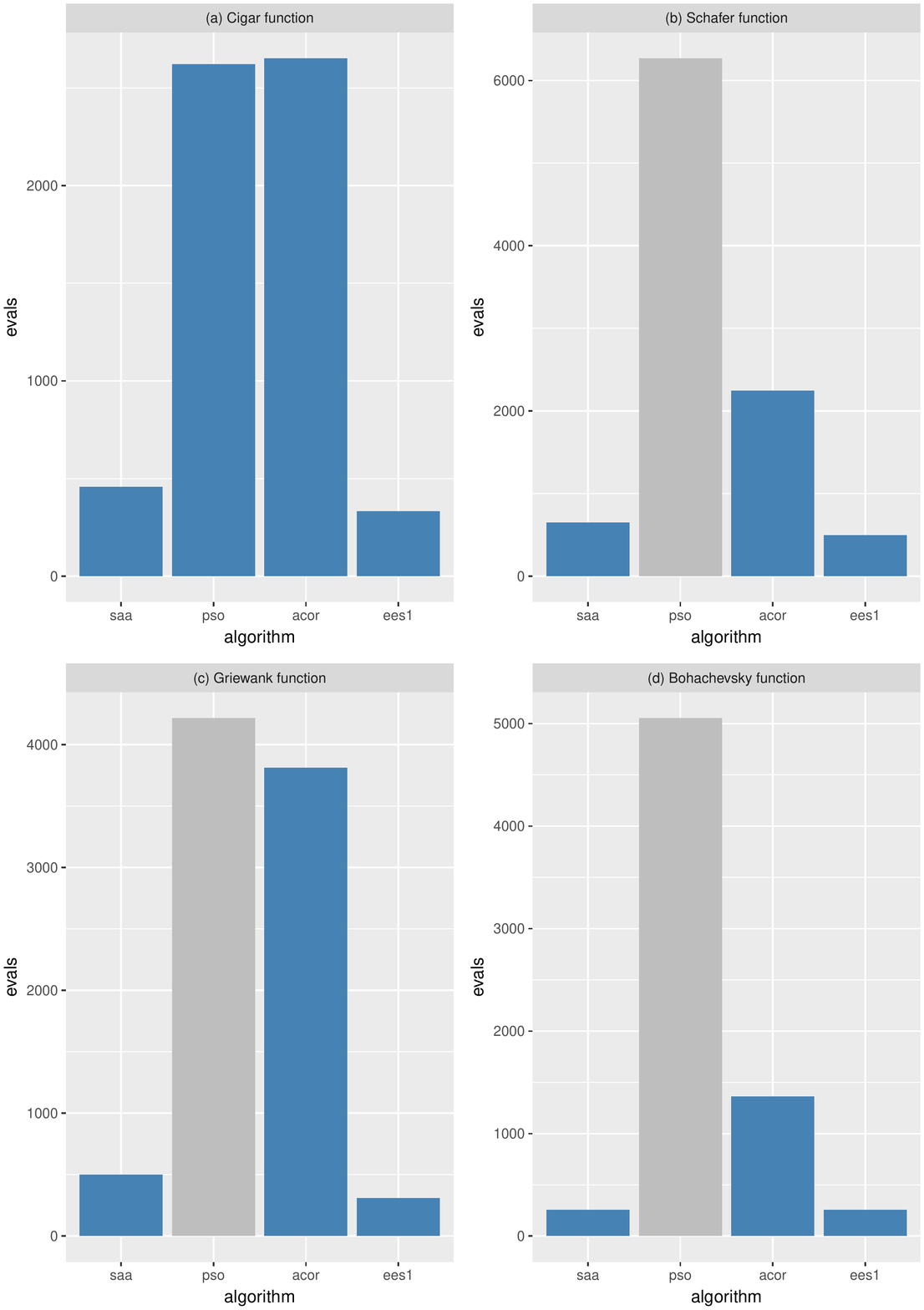}
]\caption[Comparing objective function evaluations]{\label{fig:algorithm-evals} Comparing the number of objective function evaluations required for each algorithm. The subplots show the average number of model evaluations which the metaheuristics required for reaching convergence using the benchmark functions (a) Cigar, (b) Schaffer, (c) Griewank and (d) Bohachevsky. The meaning of gray bar is that number of experiments which algorithm converge were inferior to 60\%.}
\end{figure}

The experiments were conducted setting a tolerance level of $10^{-1}$ in order to avoid a time-consuming process and for mimicking tolerance levels which may be relevant for individual-based models which may be considered to converge with high values than plain mathematical functions. These values can be taken as a starting point for deciding what algorithm is most likely to provide acceptable results for the optima with lower number of model evaluations. The Figure \ref{fig:algorithm-evals} presents graphically the results for the benchmark, showing the total number of model evaluations required for model converging with the provided tolerance level. As can be observed, the evolutionary strategy 1 (ees1) metaheuristic consistently require fewer evaluations of objective function than the other algorithms, excepting for the {\it Bohachvsky} function which required practically the same number of evaluations as the second better algorithm which is the simulated annealing (saa). The third best algorithm is the ant colony for continuous domains (acor) followed by particle swarm optimization (pso) that curiously have not behaved as expected with the parameters tuned according the recommended values \cite{Clerc2012} and we are evaluating other combination of parameters and neighborhood functions.

\begin{figure}[ht]
\centering
\includegraphics[scale=0.7]{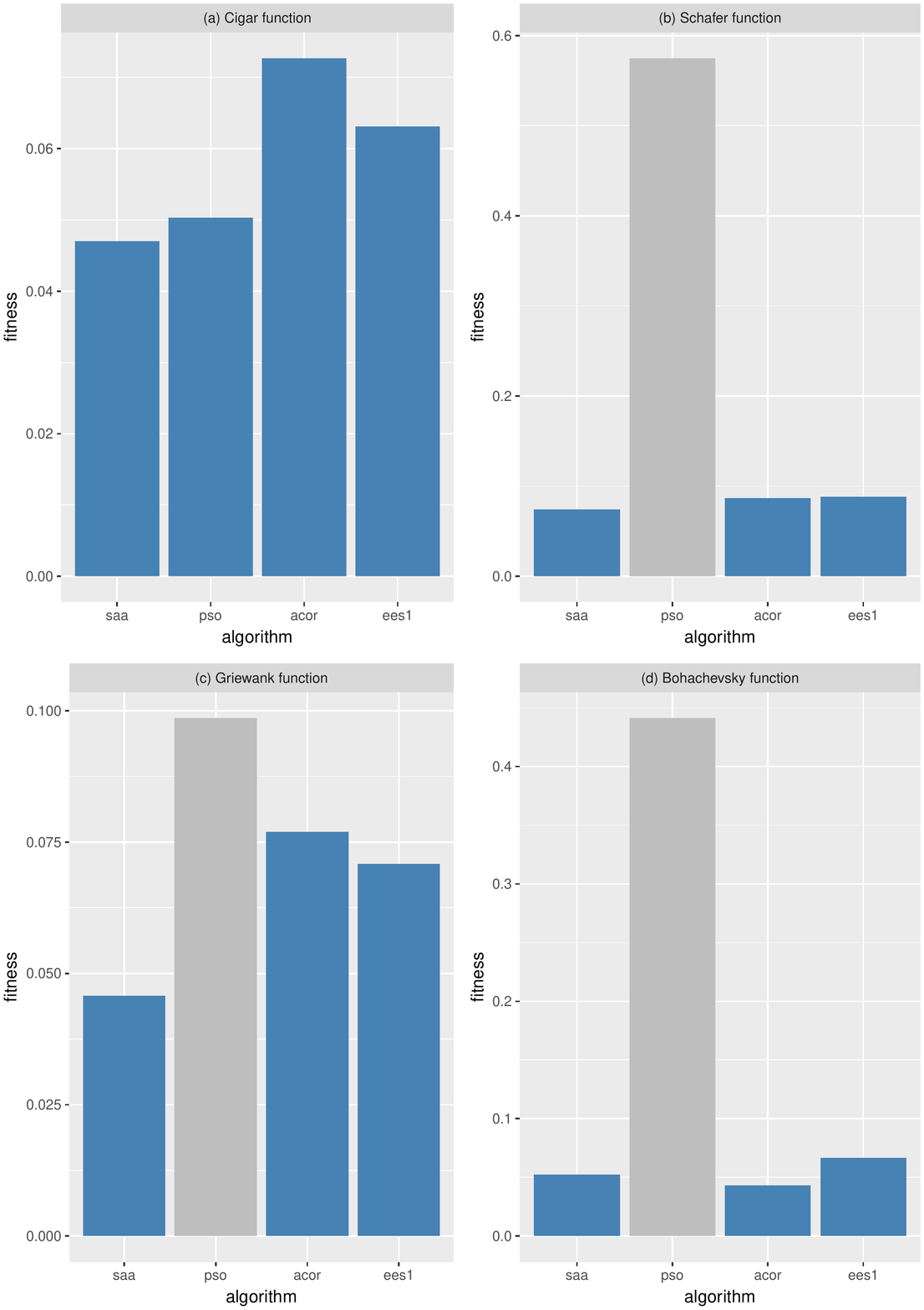}
]\caption[Comparing objective function value]{\label{fig:algorithm-fitness} Comparing the final value of objective function (fitness) for each algorithm. The subplots show the average fitness value for the benchmark functions (a) Cigar, (b) Schaffer, (c) Griewank and (d) Bohachevsky. The meaning of gray bar is that number of experiments which algorithm converge were inferior to 60\%.} 
\end{figure}

The Figure \ref{fig:algorithm-fitness} shows the value of objective function when algorithm terminates even when convergence criteria is met  or when the algorithm reaches the maximum number of configured iterations.  It is important to note that the comparisons shown here are just an initial set of hints for providing a general overview for behavior of each of the algorithms mentioned in this work. The fact of the simulated annealing has been the best performer is the expected result because differently from other algorithms, it is using a population of size $N=1$ which means that for each iteration only one individual problem solution is being evaluated. The other algorithms, by default are using values of $N=16$, $N=64$ and $N=10$ respectively the number of particles of {\it particle swarm optimization}, the number of ants of {\it ant colony optimization} and the solution size of {\it evoper evolutionary strategy 1}. This is one of the factors causing the differentiate performance figures.

Consequently, the initial parameter set, defined of each algorithm, should be seen as the starting point for tuning the metaheuristics for achieving the desired results and considerable amount of testing may be necessary for getting the best results for the parameter estimation process of a particular model instance. Additionally, some algorithm can be more adequate for a problem than another, therefore, checking the initial outputs of multiple algorithms limiting the number of iterations, may be an interesting exercise for choosing the most suitable metaheuristic. 

\subsection{Parameter estimation of individual-based models}

One of the remarkable aspects is that the syntax is simple and consistent independent of the function for which parameters are being estimated which means that the set of API primitives required for applying the metaheuristics are the same independently the target model is a plain mathematical function, an ODE, an individual-based model implemented in Repast or in any other environment for with an add-on have been implemented. The following example consists in the search for the best parameter combination which minimizes the discrepancies between the simulated values for bacterial conjugation produced using the BactoSIM simulation model \citep{PrestesGarcia2015a} \citep{PrestesGarcia2015b} and the experimental values for conjugative plasmids taken experimentally \citep{DelCampo2012}. The model provides several outputs but only two values will be used as reference for the parameter estimation process: the conjugation rates and the doubling time for donor and transconjugant cells because experimental observations are available only for these variables. Both model outputs conjugation rates and the doubling time are time series but we will just compare the first for every simulated time step and for the generation time the overall average will be employed for defining the cost metric. The conjugation rate metric used in the simulation experiments is the ration between the number of transconjugant cells and sum of transconjugant cells plus the uninfected recipients, defined as $T/(T+R)$. For building the metrics for comparing both experimental and simulated time, two approaches have been explored, one using a simple metamodeling \citep{Jin2003} \citep{Saltelli2008} using a linear model fit to the model output and the observed data and another using the dynamic time warping technique \citep{Lee2015}. These alternative approaches are presented In Figure \ref{fig:example2} where the functions {\it my.cost1} and {\it my.cost2} shows respectively the implementations of cost function using a metamodel and the dynamic time warp algorithm\citep{Giorgino2009} for comparing experimental and simulated conjugation rates time series.

\begin{figure}[h]
\begin{lstlisting}[escapeinside={(*}{*)}, basicstyle=\scriptsize, language=R]
# Step 0
rm(list=ls())
library(evoper)
set.seed(161803398)

# Step 1a
my.cost1<- function(params, results) {
   results<- results[results$Time > 60,]
   
   mm1<- with(results,lm(X.Simulated. ~ Time, data = results))
   mm2<- with(results,lm(X.Experimental. ~ Time, data = results))
   
   rate<- AoE.RMSD(coef(mm1)[1], coef(mm2)[1]) + AoE.RMSD(coef(mm1)[2], coef(mm2)[2]) 
   
   gT<- AoE.RMSD( ifelse((results$G.T. > 41 & results$G.T. < 63), 52, results$G.T.), 52 ) 
   gD<- AoE.RMSD( ifelse((results$G.D. > 32 & results$G.D. < 54), 43, results$G.T.), 43 ) 
      
   criteria<- cbind(rate,gT,gD)
   return(criteria)
}

# Step 1b
my.cost2<- function(params, results) {
   results<- results[results$Time > 60,]
   alignment<-dtw(results$X.Simulated,results$X.Experimental,keep=TRUE)
   rate<- alignment$distance
   
   (* \dots *)
   
   criteria<- cbind(rate,gT,gD)
   return(criteria)
}

# Step 2
objective<- RepastFunction$new("/usr/models/BactoSim","ds::Output",360,my.cost)

# Step 3
objective$Parameter(name="cyclePoint",min=1,max=90)
objective$Parameter(name="conjugationCost",min=0,max=100)
objective$Parameter(name="pilusExpressionCost",min=0,max=100)
objective$Parameter(name="gamma0",min=1,max=10)

# Step 4
objective$setTolerance(0.1)

my.options<- OptionsACOR$new()
my.options$setValue("iterations", 30)

# Step 5
results<- extremize("acor", objective, my.options)
\end{lstlisting}
\caption[Parameter estimation of IbM]{\label{fig:example2} The code required running the parameter estimation for an individual-based model using the Ant Colony Optimization algorithm. This code snippet shows the implementation details for two alternative implementations of cost function. As can be seen, the sequence of steps required are: {\it Step 0} loading the required libraries and sets the random seed; The {\it Step 1a} and {\it Step 1b} are the implementation of two alternative cost functions one using a metamodel fitted to the simulated data and another the dynamic time warping distance as the cost metric; {\it Step 2} Creates an instance of a {\it RepastFunction} class for the underlying model, initializing the model directory and the total simulated time; The {\it Step 3} initialize the model parameters of interest which can be a subset of all declared parameters; The {\it Step 4} shows the creation of a non-default options class for setting the maximum number of algorithm iterations to 30; Finally, in the {\it Step 5}  the {\bf extremize} function perform the optimization of cost function. This example shows how find the best combination of model parameters which minimize the differences between the observed and the simulated data for the simulated variables {\it conjugation rate} and {\it doubling time}.}
\end{figure}

The implementation of cost function shown in {\it Step 1a} creates two simple linear regression models of observed and simulated data for comparing the slope and the intercept coefficients which are serves as the distance metric for measuring how close are both time series using root mean square deviations. The cost function also considers the values of doubling time for creating a composite metric. Additionally, the function {\it my.cost1} uses a hybrid categorical-quantitative metric for the {\it doubling time} output which is described in the Equations ~\eqref{eq:cost-1}  and ~\eqref{eq:cost-2}, respectively the cost estimator for the doubling time of donors (D) and transconjugant (D) bacterial cells. Basically, the cost is zero if the values estimated by the model falls within a limited range around the average value of experimental data or the root mean square deviation (RMSD) between the simulated and observed values otherwise. 
\
\begin{equation}
\label{eq:cost-1}
\mathcal{C}(g_{D}) =	
	\begin{cases}
		0, 									& \text{if } 42 \leq g_{D} \leq 62	\\
		\text{RMSD} (g_{D},52),				& \text{otherwise}.						\\
	\end{cases}\\
\end{equation}
\
\begin{equation}
\label{eq:cost-2}
\mathcal{C}(g_{T}) =	
	\begin{cases}
		0, 									& \text{if } 33 \leq g_{T} \leq 53	\\
		\text{RMSD} (g_{T},43),				& \text{otherwise}.						\\
	\end{cases}	
\end{equation}
\

\begin{figure}[ht]
\centering
\includegraphics[scale=0.7]{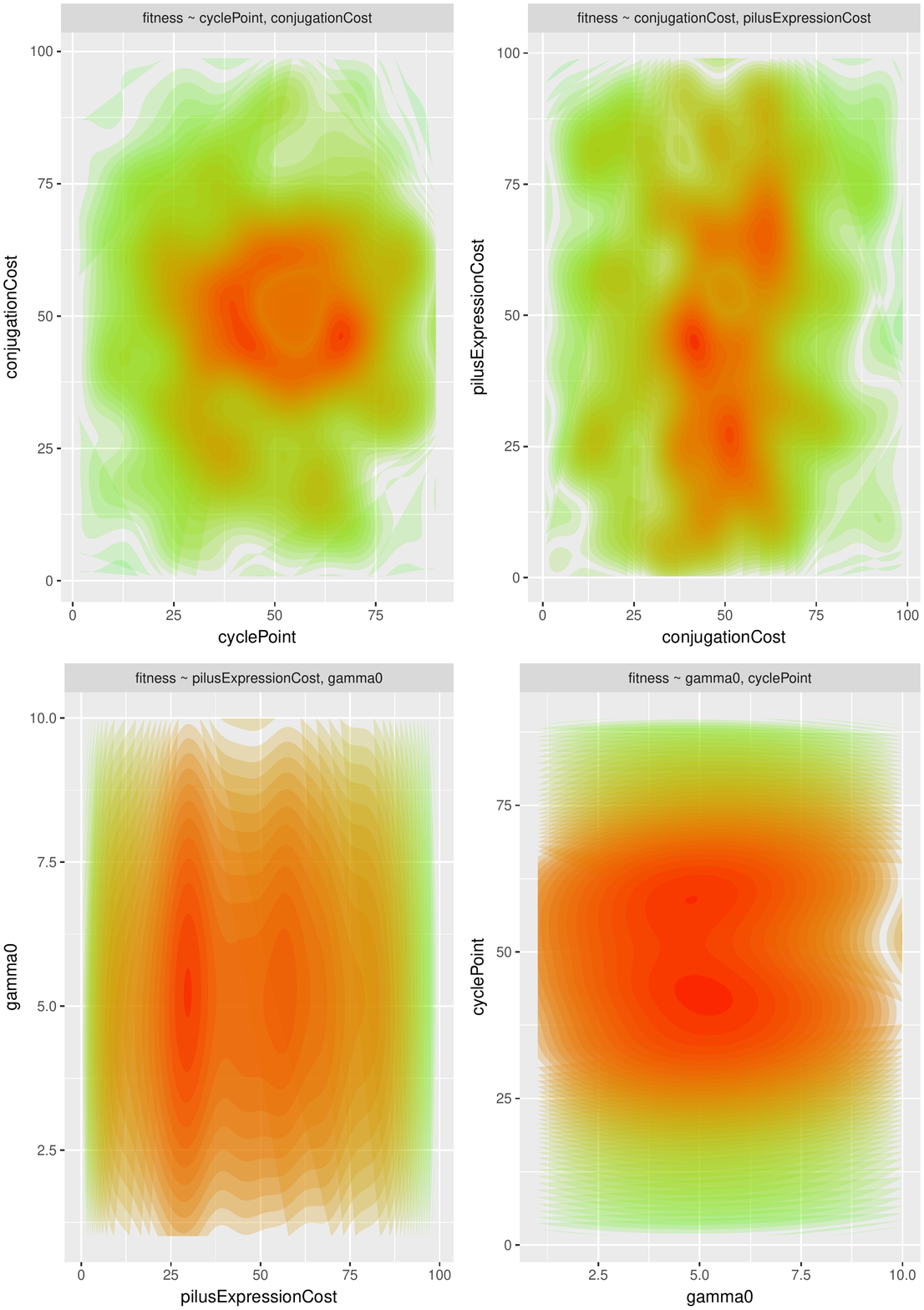}
\caption[Exploring IBM solution space]{\label{fig:ibm-exploring-1} The mapping of solution space of BactoSIM individual-based model of bacterial conjugation dynamics. The series of four contour plot shows the effects on the fitness value, defined by the cost function {\it my.cost1}, for the different values of parameter explored by the optimization algorithm.}
\end{figure}

In both cost functions {\it my.cost1} and {\it my.cost2}, the first 60 minutes are eliminated from being used in the distance metric because during that time lapse the experimental data is a somewhat noise, possibly due to the lagging time or the adaptation to the culture medium. It is also worth to mention that before undertaking a full length run, which certainly would last a large amount of time, would be much better start trying several algorithms with a limited number iterations for getting an initial map of problem solution or using the {\it ees2} metaheuristic which is limited to 600 evaluations of objective function. An example of an initial mapping for a real individual-based model is shown in Figure \ref{fig:ibm-exploring-1}. One of the things that can be perceived at a first glance is that this plot have more diffuse borders delimiting the best zones of problem solution space than the examples presented previously in Figures \ref{fig:exploring-solution-1} and \ref{fig:exploring-solution-2} where the object of study were plain mathematical functions. That is the commonly observed pattern for real models owing to the stochasticity and the nonlinear iterations between the model elements. Hence, it is normally necessary to make several initial mappings of problem for recognizing the zones of best fitness and then making the adjustments on the initial parameter ranges for achieving best estimation results.

\begin{figure}[ht]
\centering
\includegraphics[scale=0.7]{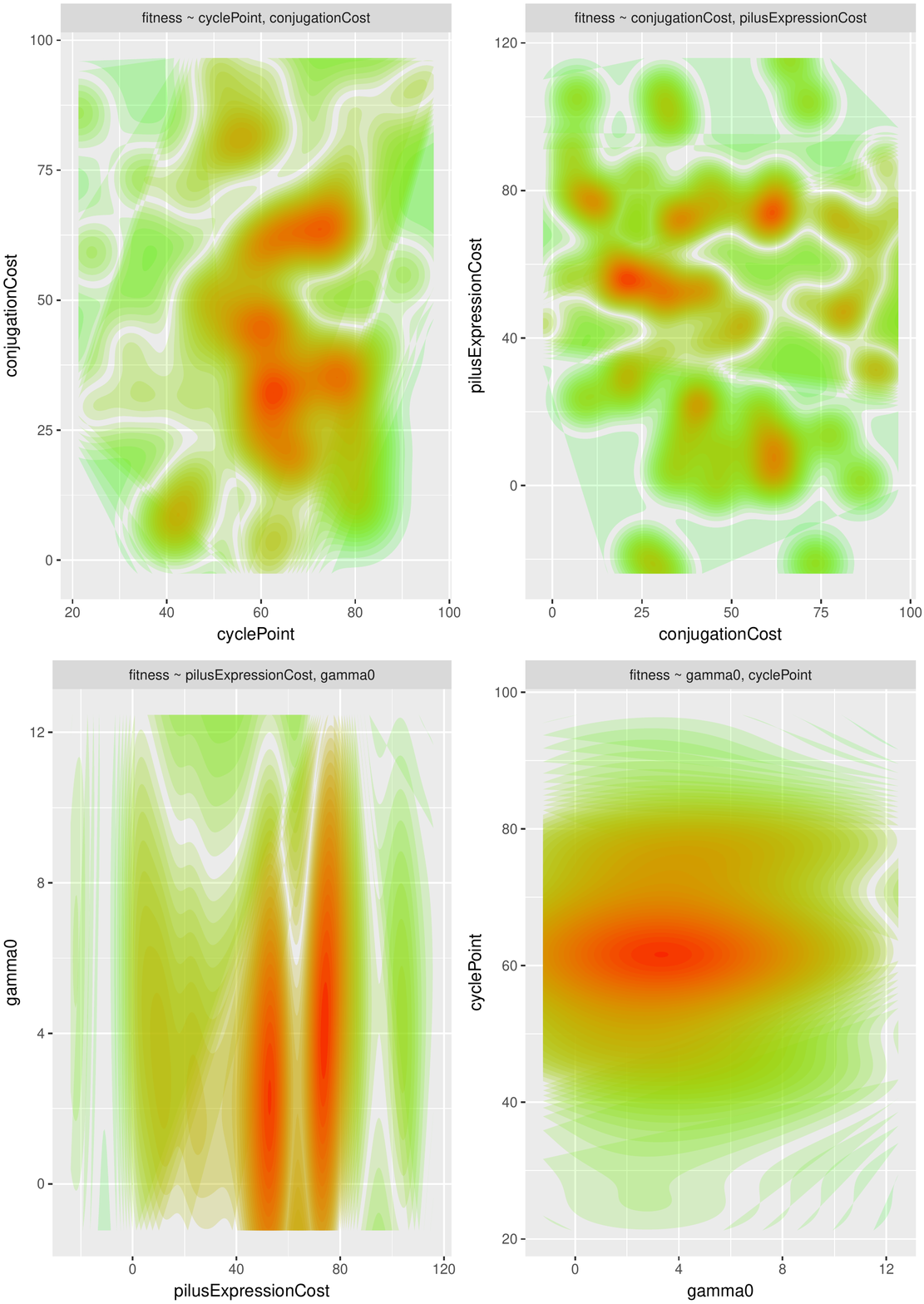}
\caption[Refining IBM solution space]{\label{fig:ibm-exploring-2} Refining the initial mapping of solution space of BactoSIM limiting the variation range of {\it cyclePoint} parameter based on the previous mapping results.}
\end{figure}

Returning to the \ref{fig:ibm-exploring-1}, despite of the lack of clearly defined limits for best parameter combinations, it can be observed that some zones are generating better cost values than others. Of course, the process should not be guided by just one algorithm with a single run but a more exhaustive exploration with several runs of available algorithms, also using different sets of random seeds. But for the sake of brevity and just illustrating the process, we will extract conclusions from this single execution.  Thus, the first contour plot which shows the parameters {\it cyclePoint}\footnote{The cycle point parameter represents the point of time, from cell birth to division when the conjugation is most likely to happen.} and {\it conjugationCost} allows to detect interesting zones for both parameters circumscribed to those values between 25\% and 75\% with peaks for {\it cyclePoint} settled approximately over values of 40\% and 70\% and the {\it conjugationCost} being close to the 50\%. The plot relating the {\it conjugationCost} and the {\it piliExpressionCost} shows similar results for the first parameter as in the previous case, which have its better fitness values nearby the 50\% of cell cycle, moreover the second has three promising zones at 25\%, 40\% and 70\%. Finally, the last two plots, relating the {\it piliExpressionCost}--{\it gamma0} and {\it gamma0}-{\it cyclePoint} are far from being conclusive but seems to indicate better performances rounding the zone of $\gamma_0 = 5$. The next step in the analysis of parameter space may include refining somehow the initial assumptions about the parameter bounds, for instance, limiting the initial range of parameter {\it cyclePoint} for the values found to contain better cost (25\%-75\%) and running again the exploratory analysis. Therefore, adjusting the range and running the {\it simulated annealing} algorithm with the maximum number of iterations limited to 50, new results are obtained and presented on Figure \ref{fig:ibm-exploring-2}. These exploratory steps can be repeated a number of times before staring a complete run of metaheuristics.

\section{Summary}

The systematic parameter estimation should be a fundamental part of individual-based modeling but it is normally omitted by modelers. One of the main reasons is the relative complexity of available methods, the effort required for applying them and the lack of simple tools for the practitioners which usually come from different domains with different backgrounds. The ecology has always greatly benefited from the application of mathematical models to the description of complex processes and iterations. Recently, the Individual-based models are becoming a lingua franca for ecological modeling but normally the acceptance of produce results are hindered by lack of a thorough parameter estimation and analysis. The cause may be attributed to the deficit of experience with methods and techniques required for carrying out the analysis of simulation output. The individual-based models are complex, stochastic and non-linear in their nature, therefore the evaluation input parameters for making the model reproducing reliably the reference data is hard computation task. The best available approach is to estate the parameter estimation as an instance of an optimization problem and apply the existing arsenal of optimization metaheuristics.

Beating this in mind, we have introduced in this work the partial set of features available on EvoPER package alongside with illustrative usage cases, including one application to a real individual-based model with the interpretation of outputs produce and the steps necessary to the complete parameter estimation of model. The package is being developed keeping in mind the idea of minimizing the effort required to the application of sophisticated methods in the parameter estimation process of Individual-based models. This package allows the modelers to try different alternatives without having to code ad hoc and complex integration code to the existent packages.

\section*{Acknowledgments}

This work was supported by the European FP7 - ICT - FET EU research project: 612146 (PLASWIRES "Plasmids as Wires" project) \url{www.plaswires.eu} and by Spanish Government (MINECO) research grant TIN2012-36992.

\bibliography{Main}

\address{Antonio Prestes Garc\'ia\\
  Departamento de Inteligencia Artificial, Universidad Polit\'ecnica de Madrid\\
  Campus de Montegancedo s/n, Boadilla del Monte, Madrid\\
  Spain\\}
\email{antonio.pgarcia@alumnos.upm.es}

\address{Alfonso Rodr\'iguez-Pat\'on\\
  Departamento de Inteligencia Artificial, Universidad Polit\'ecnica de Madrid\\
  Campus de Montegancedo s/n, Boadilla del Monte, Madrid\\
  Spain\\}
\email{arpaton@fi.upm.es}